\documentclass{article}

\usepackage[preprint]{neurips_2026}

\usepackage[utf8]{inputenc}
\usepackage[T1]{fontenc}
\usepackage{hyperref}
\usepackage{url}
\usepackage{booktabs}
\usepackage{amsfonts}
\usepackage{amsmath}
\usepackage{nicefrac}
\usepackage{microtype}
\usepackage{xcolor}
\usepackage{graphicx}
\usepackage{multirow}
\usepackage{algorithm}
\usepackage{algorithmic}
\usepackage{tikz}
\usepackage[most]{tcolorbox}
\usepackage{wrapfig}
\usepackage{pifont}
\usepackage{fontawesome5}

\definecolor{citecolor}{RGB}{66,133,244}
\hypersetup{
    colorlinks=true,
    linkcolor=citecolor,
    filecolor=citecolor,      
    urlcolor=citecolor,
    citecolor=citecolor,
}

\definecolor{scopeblue}{RGB}{30, 100, 200}
\definecolor{scopecyan}{RGB}{60, 160, 220}
\definecolor{scopeorange}{RGB}{210, 120, 40}
\definecolor{scopecopper}{RGB}{180, 80, 30}

\newcommand{\SCOPE}{%
\textbf{\textcolor{scopeblue}{S}\textcolor{scopecyan}{C}\textcolor{scopecyan}{O}\textcolor{scopeorange}{P}\textcolor{scopecopper}{E}}%
}

\definecolor{rankfirst}{RGB}{215, 240, 255}    
\definecolor{ranksecond}{RGB}{255, 235, 205}   
\definecolor{rankthird}{RGB}{240, 230, 250}    

\newtcbox{\firstbox}{
    on line, arc=2pt, outer arc=2pt,
    colback=rankfirst, colframe=rankfirst,
    boxsep=0pt, left=2pt, right=2pt, top=1.5pt, bottom=1.5pt,
    boxrule=0pt
}
\newtcbox{\secondbox}{
    on line, arc=2pt, outer arc=2pt,
    colback=ranksecond, colframe=ranksecond,
    boxsep=0pt, left=2pt, right=2pt, top=1.5pt, bottom=1.5pt,
    boxrule=0pt
}
\newtcbox{\thirdbox}{
    on line, arc=2pt, outer arc=2pt,
    colback=rankthird, colframe=rankthird,
    boxsep=0pt, left=2pt, right=2pt, top=1.5pt, bottom=1.5pt,
    boxrule=0pt
}

\newcommand{\first}[1]{\firstbox{\textbf{#1}}}
\newcommand{\second}[1]{\secondbox{#1}}
\newcommand{\third}[1]{\thirdbox{#1}}

\title{\SCOPE{}: Simulating Cross-game Operations in Playable Environments for FPS World Models}

\author{
    \textbf{Zizhao Tong}$^{1,2\dagger}$ \quad 
    \textbf{Yeying Jin}$^{2,3\ddagger\text{\ding{41}}}$ \quad
    \textbf{Hongfeng Lai}$^{2\dagger}$ \quad 
    \textbf{Zeqing Wang}$^{2,3\dagger}$ \quad
    \textbf{Zhaohu Xing}$^{4}$ \\
    \textbf{Kexu Cheng}$^{1}$ \quad
    \textbf{Haoran Xu}$^{5}$ \quad
    \textbf{Zhao Pu}$^{6}$ \quad
    \textbf{Shangwen Zhu}$^{6}$ \\
    \textbf{Ruili Feng}$^{7}$ \quad
    \textbf{Jian Zhao}$^{8}$ \quad
    \textbf{Yan Zhang}$^{3}$ \\
    \textbf{Hao Tang}$^{9}$ \quad
    \textbf{Ling Shao}$^{1\text{\ding{41}}}$ \\
    \vspace{0.3em} \\
    \parbox{\textwidth}{\centering \textnormal{
        $^1$UCAS-Terminus AI Lab, University of Chinese Academy of Sciences \quad $^2$Tencent \\
        $^3$National University of Singapore \quad $^5$Zhejiang University \quad $^6$Shanghai Jiaotong University \\
        $^4$The Hong Kong University of Science and Technology (Guangzhou) \quad $^7$University of Waterloo \\
        $^8$Zhongguancun Institute of Artificial Intelligence \\
        $^9$State Key Laboratory of Multimedia Information Processing, \\
        School of Computer Science, Peking University \\
        \vspace{0.3em}
        \texttt{tongzizhao24@mails.ucas.ac.cn \quad jinyeying@u.nus.edu \quad ling.shao@ieee.org}
    }}
}

\begin{document}

\begingroup
\renewcommand{\thefootnote}{$\dagger$} 
\footnotetext{This work was completed during a research internship at Tencent, supervised by Yeying Jin.} 
\renewcommand{\thefootnote}{$\ddagger$} 
\footnotetext{Project lead.}
\renewcommand{\thefootnote}{\ding{41}} 
\footnotetext{Corresponding Author.}
\endgroup

\maketitle

\begin{center}
    \vspace{-2.5em}
    \footnotesize
    \setlength{\tabcolsep}{0pt}
    \href{https://z2tong.github.io/SCOPE/}{\color{black}{\faGlobe}\hspace{0.2em}Project Page}
    \hspace{1.2em}
    \href{https://github.com/z2tong/SCOPE}{\color{black}{\faGithub}\hspace{0.2em}Code}
    \hspace{1.2em}
    \href{https://huggingface.co/zizhaotong/SCOPE}{\color{black}{\faGem}\hspace{0.2em}Model} 
    \hspace{1.2em}
    \href{https://huggingface.co/collections/zizhaotong/crossfps}{\color{black}{\faDatabase}\hspace{0.2em}Dataset}
    \vspace{-0.5em}
\end{center}


\begin{figure}[h]
    \centering
    \includegraphics[width=\linewidth]{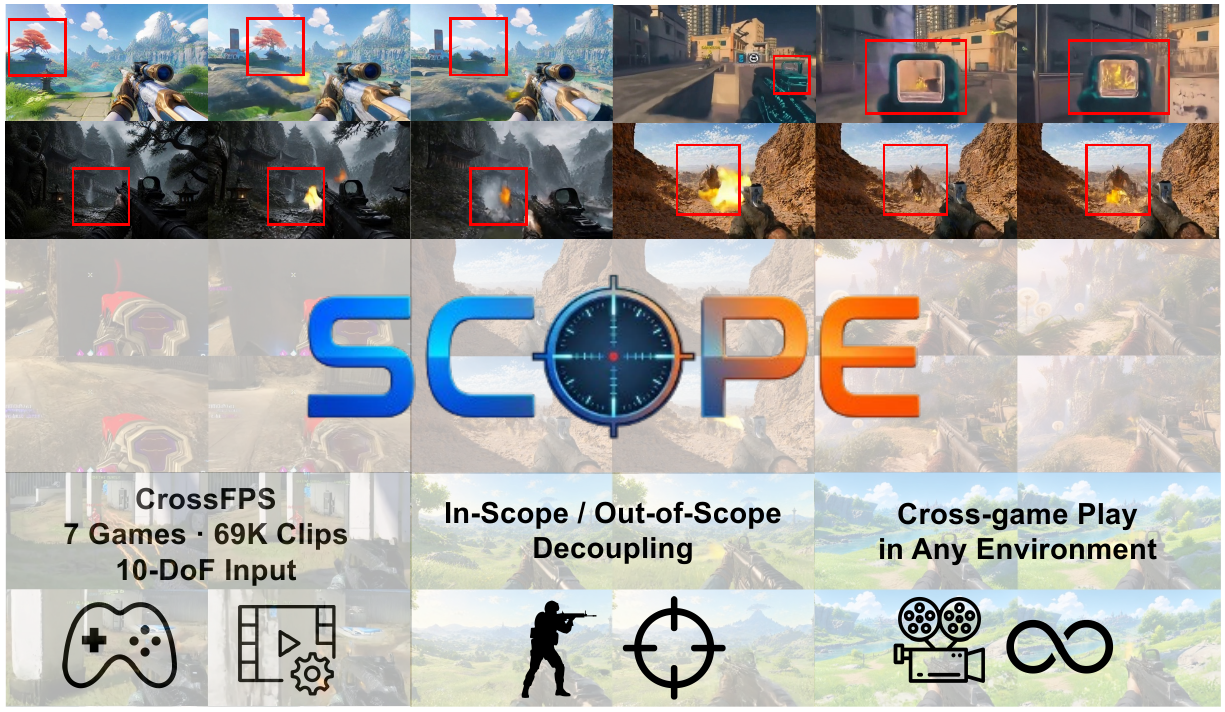}
    \caption{\SCOPE{} executes complex multi-action controls and action-environment interactions (highlighted in \textcolor{red}{red boxes}) across diverse, unseen first-person scenes without retraining.}
    \label{fig:teaser}
\end{figure}

\begin{abstract}
Interactive world models for first-person shooter (FPS) games must resolve high-frequency overlapping control signals at every frame without disrupting unaffected regions. Existing methods inject actions globally and train on single titles, failing under dense FPS inputs. We observe that FPS actions are spatially selective: discrete events such as firing or reloading affect only a localized region around the weapon (the scope), while continuous camera and movement signals govern stable surroundings. We propose \SCOPE{}, which inserts a conditioning module into each transformer block of a pretrained video diffusion model. It reshapes features into per-pixel temporal sequences so that each position computes its action response from local visual content. This separates in-scope effects from out-of-scope generation without segmentation labels. We also introduce CrossFPS, the first multi-game FPS dataset with frame-aligned action telemetry. It comprises 69K clips from 7 titles with 10-DoF controller signals, curated to remove gameplay bias. The model learns general visual-to-action mappings rather than game-specific patterns, enabling zero-shot transfer to unseen scenes. Experiments confirm strong action responsiveness, precise scope separation, and effective cross-game generalization.

\end{abstract}

\section{Introduction}
\label{sec:intro}

World models predict the consequences of actions within an environment, allowing agents to plan and interact~\citep{ha2018worldmodels,hafner2019dreamer}. Recent video diffusion models have been interpreted as implicit world simulators~\citep{nvidia2025cosmos,yang2024unisim}, enabling generative game engines that accept player inputs and produce visually coherent continuations~\citep{valevski2025gamengen,oasis,alonso2024diamond}. These systems support interactive simulation across genres from Atari to Minecraft, suggesting that video generation can serve as a general substrate for world modeling.

First-person shooter (FPS) games expose a critical failure mode of this paradigm. FPS gameplay produces exceptionally dense control signals: players execute rapid camera sweeps exceeding 180\textdegree/s, interleave simultaneous firing and movement, and chain multiple discrete events within a single generation window. Current world models inject actions through global conditioning~\citep{oasis,tang2025hunyuan,gamegenx} that broadcasts a single embedding uniformly across all spatial positions. Under sparse, low-frequency controls such as open-world navigation, global injection suffices. Under the high-frequency regime of FPS, it collapses: a firing command intended for one localized region simultaneously perturbs every pixel, and rapid successive inputs compound distortions across frames. The core issue is that global conditioning cannot distinguish where in the frame each action should take effect.

We observe that FPS actions are spatially selective. Discrete events such as firing or reloading manifest only within a localized region around the weapon and immediate interaction area, which we term the scope. Everything outside the scope, including walls, sky, and distant environment, should remain stable under continuous camera and movement controls. This suggests a natural decomposition. In-scope regions require focused modeling of discrete action-to-visual correspondences, which is easier to learn in a confined spatial context than across the entire frame. Out-of-scope regions require stable scene generation driven by continuous ego-motion, which benefits from excluding in-scope dynamics so that out-of-scope synthesis is not contaminated by localized effects. Both sides demand the same primitive: per-pixel conditioning that lets each position determine whether it lies in-scope or out-of-scope from its local visual content.

Based on this observation, we propose \SCOPE{}. This conditioning module is inserted into each transformer block of a pretrained video diffusion model. It reshapes features into per-pixel temporal sequences so that each position independently computes its action response from local visual content. Discrete events are processed via visually-queried cross-attention that confines effects to in-scope regions. Continuous controls are routed through temporal self-attention that models smooth ego-motion for out-of-scope generation. All modules are zero-initialized, so training begins from an unmodified video generator and progressively acquires scope separation without segmentation labels.

Existing game world models train on single titles~\citep{alonso2024diamond,valevski2025gamengen,oasis}, yet FPS games share common action-visual dynamics across titles: firing produces a muzzle flash, rightward aiming induces leftward scene flow. No prior dataset provides multi-game coverage with dense frame-aligned action annotation. We therefore introduce CrossFPS, comprising 69,000 clips across seven FPS titles with 10-dimensional controller telemetry, curated to remove gameplay bias. Training on CrossFPS enables the model to learn general visual-to-action mappings rather than game-specific patterns, allowing zero-shot transfer to unseen scenes without retraining.

Our contributions are threefold. We propose SCOPE, whose per-pixel conditioning decomposes action effects into in-scope discrete responses and out-of-scope continuous generation through end-to-end training without segmentation supervision. We introduce CrossFPS, the first multi-game FPS dataset with frame-aligned action telemetry. We demonstrate robust controllability on unseen scenes, effective zero-shot generalization, and evidence that the architecture benefits from data scaling.

\section{Related Work}
\paragraph{World Models.}
World models learn environment dynamics to support prediction, planning, and control~\citep{craik1943nature,ha2018worldmodels,ding2024survey_wm,chu2026agentic}. In reinforcement learning, they simulate transition dynamics before execution~\citep{sutton1991dyna,hafner2023dreamerv3,schrittwieser2020muzero}. In computer vision, world models typically manifest as video generators that produce temporally coherent continuations~\citep{brooks2024sora,bruce2024genie,nvidia2025cosmos}. A growing body of literature further pursues long-horizon consistency~\citep{yu2025context,xiao2025worldmem,nam2026worldcam,sun2025worldplay}, long-horizon memory~\citep{wang2026matrix}, physical plausibility~\citep{wang2025wisa}, and real-time inference~\citep{yin2023one0step,zhu2026causal,zhu2026sana}. The unifying principle is that agents rely on internal models to anticipate the outcomes of actions, whether for policy optimization in simulation or for interactive content generation. Our work falls into this category; we develop an interactive world model that conditions video generation on dense player actions, specifically maintaining structural consistency under complex, high-frequency control signals, ensuring stable frame transitions during gameplay.
\paragraph{Video Diffusion Models.}
Diffusion-based generative models~\citep{ddpm,sde,score} have driven rapid progress in visual synthesis. In the image domain, latent diffusion~\citep{sd} and its successors~\citep{pixart,sdxl} produce high-fidelity outputs at scale. In the video domain, frameworks such as VideoCrafter~\citep{videocrafter2}, SVD~\citep{svd}, Open-Sora~\citep{opensora,open-sora-plan}, CogVideoX~\citep{cogvideox}, HunyuanVideo~\citep{kong2024hunyuanvideo}, and Wan~\citep{wan} achieve temporally coherent generation across diverse content. The transition to Transformer-based architectures~\citep{dit,brooks2024sora} has further improved generation quality and scalability, leading researchers to interpret video diffusion models as implicit physical simulators~\citep{yang2024unisim,nvidia2025cosmos} with applications in autonomous driving~\citep{min2024driveworld} and robotics~\citep{wu2023daydreamer}. Our work builds on this foundation by extending a pretrained video DiT into an interactive world model via per-pixel action conditioning, successfully mapping fine-grained input sequences to specific visual changes instead of relying on global representations.
\paragraph{Game World Models.}
Games provide natural test beds for interactive world models due to the combination of visual dynamics and rule-based logic~\citep{ding2024survey_wm}. Early GAN-based methods~\citep{drivegan,gamegan} demonstrated limited generative capabilities. Subsequent diffusion-based systems~\citep{bruce2024genie,parker2024genie2,genie3,alonso2024diamond} have considerably advanced interactive video generation~\citep{IGVsurvey}, enabling world models for specific titles such as Atari~\citep{alonso2024diamond}, DOOM~\citep{valevski2025gamengen}, and Minecraft~\citep{oasis,guo2025mineworld}. However, existing methods are often constrained by simplified action spaces, relying on sparse discrete keystrokes~\citep{genie3,valevski2025gamengen}, low-dimensional continuous controls~\citep{team2026advancing}, or coarse text instructions~\citep{gamegenx} that fail to capture instantaneous inputs. Furthermore, injecting actions through global mechanisms, such as adaptive normalization~\citep{oasis,tang2025hunyuan}, cross-attention tokens~\citep{gamegenx}, or latent action codes~\citep{bruce2024genie,alonso2024diamond}, broadcasts a uniform action signal to all spatial positions. This conflates in-scope regions that require localized animation with out-of-scope regions that should remain stable, a mismatch that worsens under the dense, high-frequency controls of First-Person Shooter gameplay. Crucially, standard world models lack action compositionality and struggle with the simultaneous execution of hybrid controls, often causing structural artifacts or total responsiveness collapse under overlapping inputs. While certain scale-oriented models pursue cross-game generalization~\citep{parker2024genie2,genie3,team2026advancing,gamefactory}, they require immense proprietary datasets or degrade when transferred to unseen domains with high-frequency control. In contrast, our approach, SCOPE, supports a comprehensive hybrid action space with dense, high-frequency control. By learning spatially selective action conditioning rather than expanding data volume, SCOPE achieves robust action composition and excels in zero-shot cross-game generalization across diverse environments using a compact 69K-clip dataset, establishing a highly scalable open-world simulation framework.

\begin{figure*}[t]
    \centering
    \includegraphics[width=\linewidth]{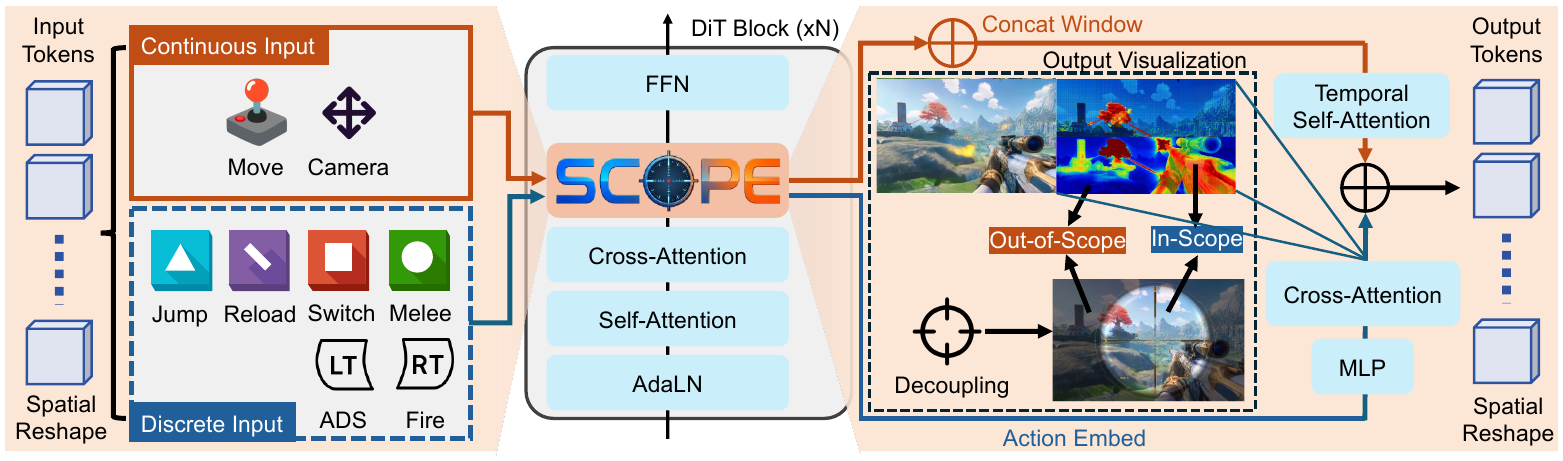}
    \caption{\SCOPE{} architecture. A SCOPE module is inserted into each DiT block. Discrete inputs use cross-attention with visual queries to confine effects to in-scope regions. Continuous inputs use MLP fusion and temporal self-attention for out-of-scope generation. Pathways combine via residual connections.}
    \label{fig:method}
\end{figure*}

\section{Method}
\label{sec:method}

\subsection{Overview}
\label{sec:overview}

Given an initial frame $I_1$ and a sequence of player actions $\mathbf{a}_{1:T}$ comprising continuous analog controls (camera, movement) and discrete button events (fire, reload, etc.), the model generates a video continuation $V_{2:T}$ that faithfully reflects the specified controls. This requires causal conditioning: each frame $V_t$ must respond to the concurrent action $\mathbf{a}_t$ rather than merely extrapolating visual momentum. As established in Section~\ref{sec:intro}, FPS actions produce spatially heterogeneous effects: discrete events should animate only in-scope regions, while continuous controls drive stable out-of-scope generation. Global action injection cannot provide this distinction.

Our method addresses this by inserting a SCOPE module into each transformer block of a pretrained video diffusion model (Figure~\ref{fig:method}). The module reshapes features into per-pixel temporal sequences and routes discrete events and continuous controls through dedicated attention pathways. Discrete events are handled via visually-queried cross-attention that confines effects to in-scope regions. Continuous controls are handled via temporal self-attention for smooth out-of-scope ego-motion. All output projections are zero-initialized so that training begins from an unmodified video generator. The entire model is trained end-to-end on CrossFPS with a flow matching objective and stochastic action dropout for Action Classifier-Free Guidance (Action-CFG) at inference.

\subsection{Preliminaries}
\label{sec:preliminary}

The model builds on a pretrained video Diffusion Transformer (DiT)~\citep{dit} with approximately five billion parameters. A 3D VAE encoder compresses input video $\mathbf{V} \in \mathbb{R}^{3 \times T \times H \times W}$ into latent representations $\mathbf{z} \in \mathbb{R}^{C \times f \times h \times w}$, where $f$, $h$, $w$ denote the compressed temporal, height, and width dimensions (temporal compression ratio 4, spatial compression ratio 8). The latents are patchified into a token sequence $\mathbf{x} \in \mathbb{R}^{B \times N \times D}$, where $B$ is the batch size, $N = f \times h \times w$ is the number of tokens, and $D$ is the hidden dimension. The backbone consists of $L=30$ transformer blocks, each containing AdaLN, self-attention with 3D RoPE, text cross-attention, and a FFN.

We adopt flow matching~\citep{lipman2023flow} as the training framework. Given clean latents $\mathbf{z}_0$ and Gaussian noise $\boldsymbol{\epsilon} \sim \mathcal{N}(0, \mathbf{I})$, noisy latents are constructed as $\mathbf{z}_t = (1 - t) \mathbf{z}_0 + t \boldsymbol{\epsilon}$ for timestep $t \in [0, 1]$. The model learns to predict the velocity field $\mathbf{v}_\theta(\mathbf{z}_t, t, \mathbf{c})$ by minimizing:
\begin{equation}
    \mathcal{L} = \mathbb{E}_{t, \mathbf{z}_0, \boldsymbol{\epsilon}} \left[ w(t) \left\| \mathbf{v}_\theta(\mathbf{z}_t, t, \mathbf{c}) - (\boldsymbol{\epsilon} - \mathbf{z}_0) \right\|^2 \right],
\end{equation}
where $\mathbf{c}$ denotes conditioning signals (text, first frame) and $w(t)$ is a timestep-dependent weight. Following the image-to-video paradigm, the first-frame latent replaces the noisy latent at the first temporal position, and the loss is computed only over subsequent frames. This formulation provides a natural foundation for action-conditioned generation: we extend $\mathbf{c}$ to include player actions via the SCOPE module described below.

\subsection{SCOPE Module}
\label{sec:action_module}

The SCOPE module is inserted between text cross-attention and FFN in each of the $L=30$ transformer blocks. It re-routes action conditioning through per-pixel temporal sequences so that each spatial location accumulates only action information relevant to its local visual content.

\paragraph{Action Representation.} FPS gameplay produces two categories of control signals (Figure~\ref{fig:method}, left). Continuous controls $\mathbf{a}_c \in \mathbb{R}^{T_{\mathrm{raw}} \times d_c}$ are captured from analog sticks, where $T_{\mathrm{raw}}$ is the number of raw gameplay frames and $d_c = 4$ covers the two movement axes and two camera axes. Discrete events $\mathbf{a}_d \in \mathbb{R}^{T_{\mathrm{raw}} \times d_d}$ are captured from button presses, where $d_d = 6$ covers fire, ADS, reload, jump, melee, and weapon switch.

\paragraph{Spatial Reshape.} The visual effect of any action depends on spatial content: identical inputs should produce different responses at different positions. To enable per-pixel conditioning, we reshape the token sequence $\mathbf{x}$ into per-pixel temporal sequences:
\begin{equation}
    \mathbf{x} \in \mathbb{R}^{B \times (f \cdot h \cdot w) \times D} \longrightarrow \hat{\mathbf{x}} \in \mathbb{R}^{(B \cdot h \cdot w) \times f \times D},
\end{equation}
where each of the $h \cdot w$ spatial positions now holds an independent temporal sequence of length $f$. All subsequent processing operates on these per-pixel sequences $\hat{\mathbf{x}}$, ensuring that in-scope and out-of-scope pixels respond differently to the same control inputs.

\paragraph{Dual-Pathway Processing.} The two action categories are processed through dedicated pathways (Figure~\ref{fig:method}).

Discrete events trigger instantaneous, spatially localized effects: firing produces a muzzle flash, scoping triggers zoom, interactions cause localized reactions. The discrete signal $\mathbf{a}_d$ is first embedded into action tokens via an MLP, then processed through cross-attention where the per-pixel features $\hat{\mathbf{x}}$ serve as queries and the action embeddings serve as keys and values:
\begin{equation}
    \Delta \mathbf{x}_d = \mathrm{CrossAttn}\!\left(Q{=}\hat{\mathbf{x}},\; K{=}V{=}\mathrm{MLP}_{\mathrm{embed}}(\mathbf{a}_d)\right).
\end{equation}
The output $\Delta \mathbf{x}_d$ represents per-pixel discrete action residuals. Since queries derive from local visual content, in-scope pixels attend strongly to action signals while out-of-scope pixels produce near-zero attention, confining discrete effects to relevant spatial regions. This mechanism requires no explicit region annotations; the separation emerges naturally from the visual content itself during training.

Continuous controls drive smooth ego-motion that primarily affects out-of-scope regions (scene flow from camera rotation, parallax from movement). For each latent frame $i$, we extract a temporal window $\mathbf{w}_i = \mathbf{a}_c[i \cdot r : i \cdot r + r \cdot s]$ of raw-frame actions, where $r=4$ is the temporal compression ratio and $s$ is the window size. This window is flattened and concatenated with the per-pixel feature $\hat{\mathbf{x}}$, then processed through a fusion MLP followed by temporal self-attention with RoPE:
\begin{equation}
    \tilde{\mathbf{x}} = \mathrm{MLP}_{\mathrm{fuse}}([\hat{\mathbf{x}} \,;\, \mathrm{flatten}(\mathbf{w})]), \quad \Delta \mathbf{x}_c = \mathrm{SelfAttn}(\tilde{\mathbf{x}}, \mathrm{RoPE}_t).
\end{equation}
The output $\Delta \mathbf{x}_c$ represents per-pixel continuous action residuals. Because the discrete pathway already captures in-scope dynamics, the continuous pathway focuses on stable out-of-scope generation without contamination from localized effects.

The two residuals are combined and added back to the original features ($\hat{\mathbf{x}} + \Delta \mathbf{x}_c + \Delta \mathbf{x}_d$), then reshaped to the standard token layout before entering the FFN.

\subsection{Training and Inference}
\label{sec:training}

The pretrained backbone and all $L=30$ SCOPE modules are trained end-to-end on CrossFPS. All SCOPE output projections are zero-initialized so the model starts as an unmodified video generator and progressively learns action conditioning. This ensures training stability while enabling the backbone to co-adapt its internal representations with the action pathways. End-to-end training yields substantially stronger results than frozen or two-stage alternatives (Section~\ref{sec:ablation}). Training uses balanced sampling across all seven titles to prevent single-source dominance. The SCOPE module adds minimal parameters relative to the backbone and operates independently per spatial position, so the architecture scales naturally with larger backbones and more training data without architectural modification.

To enable tunable action intensity at inference, we apply stochastic action dropout during training: with probability $p_{\mathrm{drop}}$, all action inputs $(\mathbf{a}_c, \mathbf{a}_d)$ are replaced by a learnable null embedding $\mathbf{a}_{\mathrm{null}}$. At inference, Action-CFG interpolates between the conditional and unconditional velocity predictions:
\begin{equation}
    \hat{\mathbf{v}} = \mathbf{v}_\theta(\mathbf{z}_t, \mathbf{a}_{\mathrm{null}}) + \lambda \left[ \mathbf{v}_\theta(\mathbf{z}_t, \mathbf{a}_c, \mathbf{a}_d) - \mathbf{v}_\theta(\mathbf{z}_t, \mathbf{a}_{\mathrm{null}}) \right],
\end{equation}
where the guidance scale $\lambda > 0$ controls action intensity ($\lambda{=}1$: standard conditioning; $\lambda{>}1$: amplified response; $\lambda{<}1$: attenuated response). Full pseudocode is provided in Appendix~\ref{app:implementation}.

\begin{figure*}[t]
    \centering
    \includegraphics[width=\linewidth]{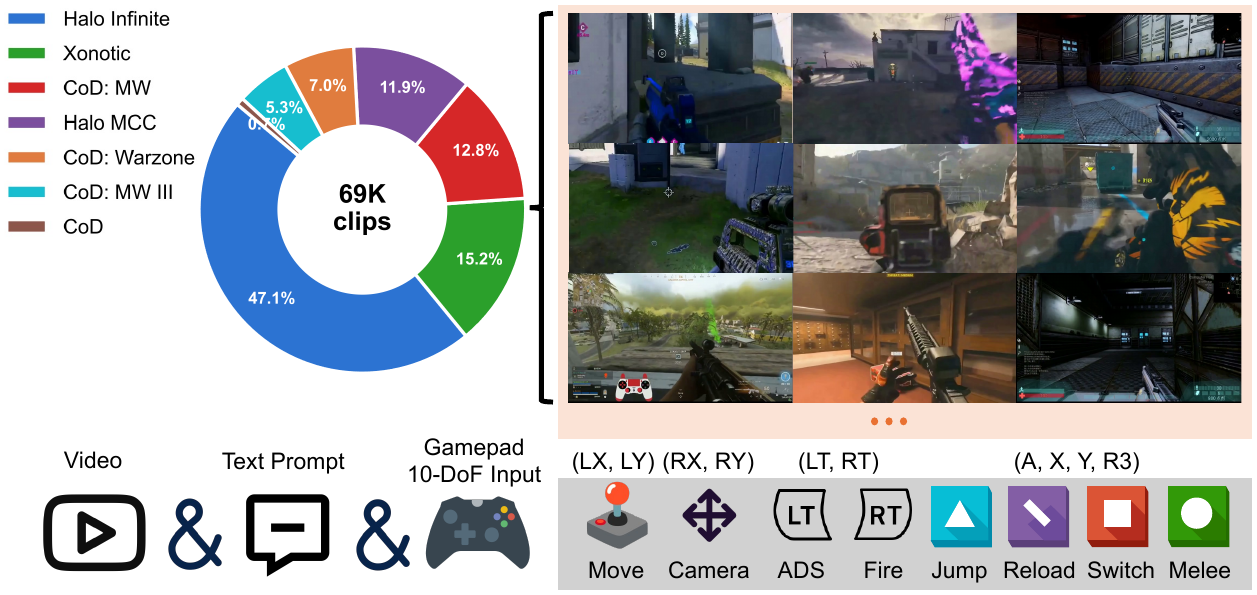}
    \caption{CrossFPS overview. Clip distribution across 7 FPS titles (69K total) with frame-aligned 10-DoF gamepad telemetry.}
    \label{fig:dataset_overview}
\end{figure*}

\section{Experiments}
\label{sec:experiments}

We evaluate our method through quantitative comparison with baselines (Section~\ref{sec:quantitative}), ablation studies (Section~\ref{sec:ablation}), and zero-shot generalization to unseen scenes (Section~\ref{sec:action_control}).

\subsection{Setup}

\paragraph{Pretrained model.}
The model builds on Wan2.2-TI2V-5B~\citep{wan}, a 5B-parameter video diffusion transformer with temporal compression ratio $r{=}4$ and spatial compression ratio 8.

\paragraph{Training.}
The backbone and 30 SCOPE modules are trained end-to-end with zero-initialized output projections. We use $480{\times}832$ resolution, 81 frames per clip (5s at 20fps), Adam with learning rate $1{\times}10^{-5}$, action dropout $p_{\mathrm{drop}}{=}0.1$, and balanced game sampling. Training takes approximately 18 hours on 8 NVIDIA GPUs.

\paragraph{CrossFPS dataset.}
CrossFPS contains 69,000 five-second clips across seven FPS titles at 20fps ($480{\times}832$), sourced from NitroGen~\citep{magne2026nitrogen} and WorldCam~\citep{nam2026worldcam}. Each clip is paired with frame-aligned 10-dimensional controller telemetry (4 continuous axes + 6 discrete buttons). The dataset is split 95:3:2 into train/val/test (65,557/2,065/1,378). Three curation stages ensure cross-game consistency: Action Distribution Balancing oversamples high-intensity clips to counteract long-tail dominance; Visual-Action De-biasing retains clips with low scene-action mutual information to prevent learning game strategies; Kinetic Normalization applies optical flow-based gain calibration to align action-to-pixel-displacement ratios across titles ($\sigma^2_{\mathrm{gain}} = 0.034$ post-normalization). Key statistics are shown in Figure~\ref{fig:dataset_overview}; full details in Appendix~\ref{app:dataset}.

\paragraph{Metrics.}
We measure action responsiveness via Dynamic Degree~\citep{huang2024vbench} and Flow Score~\citep{liu2024evalcrafter}; spatial stability via Photometric Smoothness~\citep{duan2025worldscore} and Depth Accuracy~\citep{shang2026worldarena}; visual quality via JEPA Similarity~\citep{bardes2024vjepa,luo2024beyond}, FVD~\citep{unterthiner2018towards}, LPIPS~\citep{zhang2018unreasonable}, and Motion Smoothness~\citep{duan2025worldscore,zhang2024vfimamba}. Computation details are in Appendix~\ref{app:metrics}. In all tables, results are highlighted as \first{first}, \second{second}, and \third{third}.

\paragraph{Baselines.}
We compare against three state-of-the-art interactive world models that support action-conditioned generation: Matrix-Game 3.0~\citep{wang2026matrix}, LingBot-World (Act)~\citep{team2026advancing}, and HY-World 1.5~\citep{tang2025hunyuan}. All three accept action signals as input but use global conditioning mechanisms. Since their native action interfaces differ from our 10-DoF telemetry format, we use Gemini~\citep{team2023gemini} to translate our action sequences into the detailed natural language prompts each baseline expects.

\begin{figure*}[t]
    \centering
    \includegraphics[width=\linewidth]{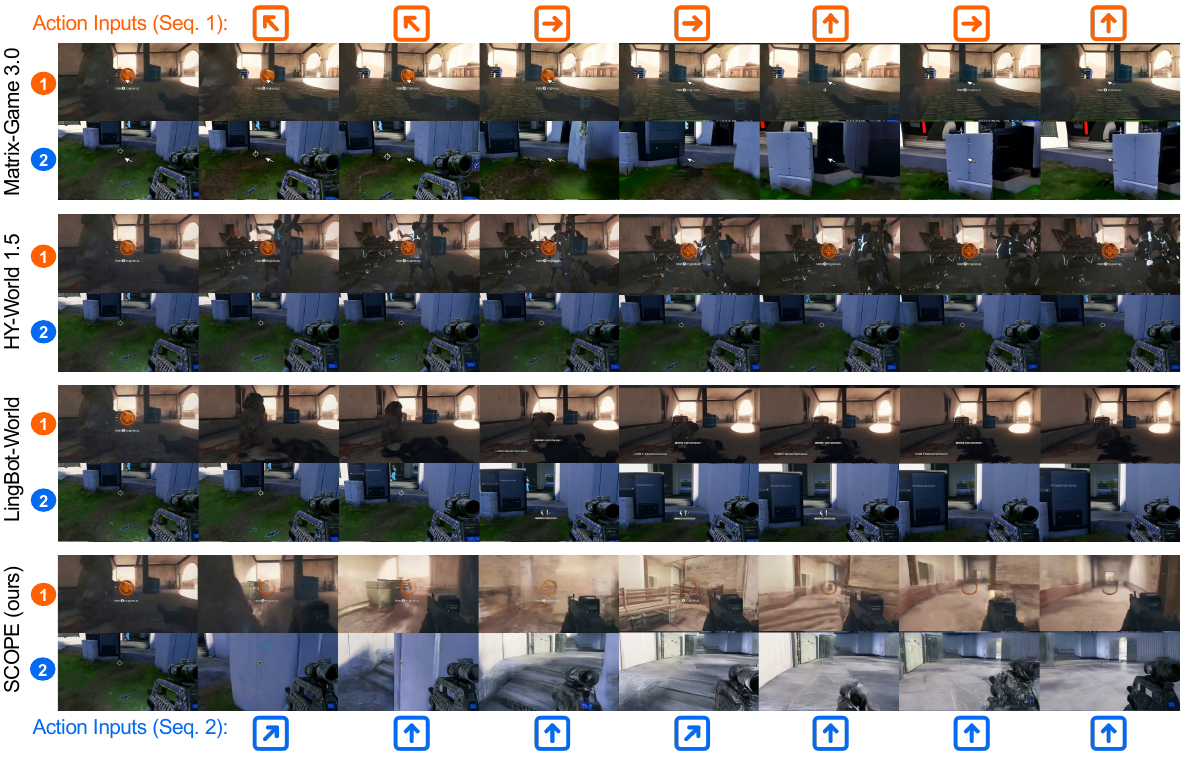}
    \caption{Qualitative comparison under high-frequency actions. Our method maintains out-of-scope stability while baselines exhibit suppressed motion, near-static output, or artifacts.}
    \label{fig:compare}
\end{figure*}

\subsection{Quantitative Comparison}
\label{sec:quantitative}

\begin{table*}[t]
\centering
\caption{Quantitative comparison on the CrossFPS test set.}
\label{tab:main_results}
\small
\resizebox{\textwidth}{!}{%
\begin{tabular}{lcccccccc}
\toprule
\multirow{2}{*}{Method} & \multicolumn{3}{c}{Visual Quality} & \multicolumn{3}{c}{Motion Quality} & \multicolumn{2}{c}{Consistency} \\
\cmidrule(lr){2-4} \cmidrule(lr){5-7} \cmidrule(lr){8-9}
& JEPA$\uparrow$ & FVD$\downarrow$ & LPIPS$\downarrow$ & Dyn.Deg.$\uparrow$ & Flow$\uparrow$ & Smooth$\uparrow$ & Photo.$\downarrow$ & Depth$\downarrow$ \\
\midrule
Matrix-Game 3.0 & 0.366 & 1022.7 & 0.692 & \third{0.661} & \third{13.36} & \first{2.502} & 1.194 & 1.524 \\
LingBot-World (Act) & \second{0.615} & \second{954.4} & \third{0.627} & \second{0.868} & \second{15.50} & \third{2.215} & \second{0.626} & \second{1.454} \\
HY-World 1.5 & \third{0.464} & 1131.7 & \second{0.611} & 0.225 & 2.37 & 1.690 & 2.523 & \third{1.502} \\
\midrule
\SCOPE{} & \first{0.806} & \first{690.3} & \first{0.601} & \first{0.910} & \first{18.24} & \second{2.383} & \first{0.198} & \first{1.299} \\
\bottomrule
\end{tabular}%
}
\end{table*}

Table~\ref{tab:main_results} shows that our method achieves the best performance on 7 of 8 metrics. The sole exception is Motion Smoothness, where Matrix-Game 3.0 leads due to action suppression rather than faithful rendering. This trade-off is expected: suppressing action responses trivially yields smoother outputs but fails the primary goal of controllability. Figure~\ref{fig:compare} confirms this qualitatively: given identical high-frequency camera rotations, our method produces smooth viewpoint changes while baselines suppress motion or introduce distortions.

The baselines receive actions through Gemini text translation rather than native telemetry, introducing an information bottleneck. To control for this modality difference, we note that the ``w/o Spatial Selectivity'' ablation in Table~\ref{tab:ablation_arch} uses native telemetry but replaces per-pixel conditioning with global injection, serving as a fair architectural comparison under identical input conditions. Its severe degradation (FVD 690.3$\to$885.4, Photo.\ 0.198$\to$0.745) confirms that the contribution is architectural rather than input-modality-driven.

Our method achieves Dynamic Degree 0.910 and Flow Score 18.24, substantially outperforming all baselines in action responsiveness. HY-World 1.5 collapses to near-static output (Dyn.Deg.\ 0.225) because its global normalization dilutes dense FPS signals below the effective threshold. Matrix-Game 3.0 attains moderate motion (0.661) but sacrifices responsiveness for smoothness. LingBot-World (0.868) performs best among baselines but loses discrete events unrecoverable from pose estimation alone. For spatial stability, Photometric Smoothness of 0.198 is $3.2\times$ better than LingBot-World (0.626) and $12.7\times$ better than HY-World (2.523), confirming scope separation without segmentation supervision. For visual quality, JEPA 0.806 (+31\% over LingBot-World), FVD 690.3 (28\% reduction), and LPIPS 0.601 (best) confirm that the model preserves backbone generation capability in out-of-scope regions while enabling precise in-scope responses.

\subsection{Ablation Studies}
\label{sec:ablation}

All ablation variants are trained identically on the full CrossFPS dataset. Results are in Table~\ref{tab:ablation_arch}.

\begin{table*}[t]
\centering
\caption{Architecture ablation on the CrossFPS test set.}
\label{tab:ablation_arch}
\small
\resizebox{\textwidth}{!}{%
\begin{tabular}{lcccccccc}
\toprule
\multirow{2}{*}{Variant} & \multicolumn{3}{c}{Visual Quality} & \multicolumn{3}{c}{Motion Quality} & \multicolumn{2}{c}{Consistency} \\
\cmidrule(lr){2-4} \cmidrule(lr){5-7} \cmidrule(lr){8-9}
& JEPA$\uparrow$ & FVD$\downarrow$ & LPIPS$\downarrow$ & Dyn.Deg.$\uparrow$ & Flow$\uparrow$ & Smooth$\uparrow$ & Photo.$\downarrow$ & Depth$\downarrow$ \\
\midrule
\SCOPE{} & \first{0.806} & \first{690.3} & \first{0.601} & \first{0.910} & \first{18.24} & \third{2.383} & \first{0.198} & \first{1.299} \\
\midrule
w/o Spatial Selectivity & 0.625 & 885.4 & 0.648 & 0.521 & 14.10 & 2.012 & 0.745 & 1.620 \\
w/o Temporal Self-Attn & 0.683 & 784.4 & 0.627 & 0.642 & 11.60 & 1.799 & 0.482 & 1.521 \\
w/o Discrete Cross-Attn & \second{0.763} & \second{725.3} & \second{0.606} & \second{0.846} & \second{17.14} & \first{2.442} & \third{0.234} & \second{1.334} \\
w/o Action-CFG & \third{0.740} & \third{725.8} & \third{0.610} & \third{0.820} & \third{15.90} & \second{2.405} & \second{0.280} & \third{1.350} \\
\midrule
\multicolumn{9}{l}{\textit{Training strategy variants}} \\
Frozen backbone & 0.724 & 775.4 & 0.631 & 0.796 & 15.57 & 2.335 & 0.264 & 1.392 \\
Two-stage (FT $\to$ freeze) & 0.761 & 732.1 & 0.614 & 0.852 & 17.13 & 2.374 & 0.226 & 1.337 \\
\bottomrule
\end{tabular}%
}
\end{table*}

\begin{figure*}[t]
\centering
\begin{minipage}[t]{0.48\linewidth}
    \centering
    \includegraphics[width=\linewidth]{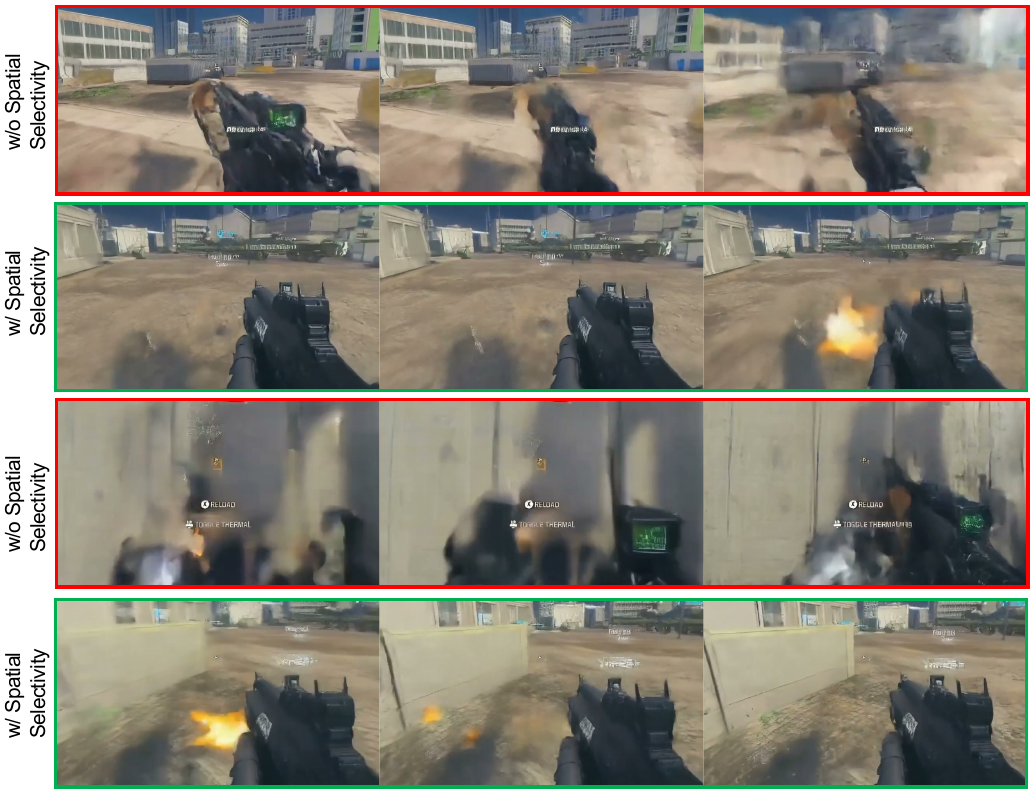}
\end{minipage}%
\hfill
\begin{minipage}[t]{0.48\linewidth}
    \centering
    \includegraphics[width=\linewidth]{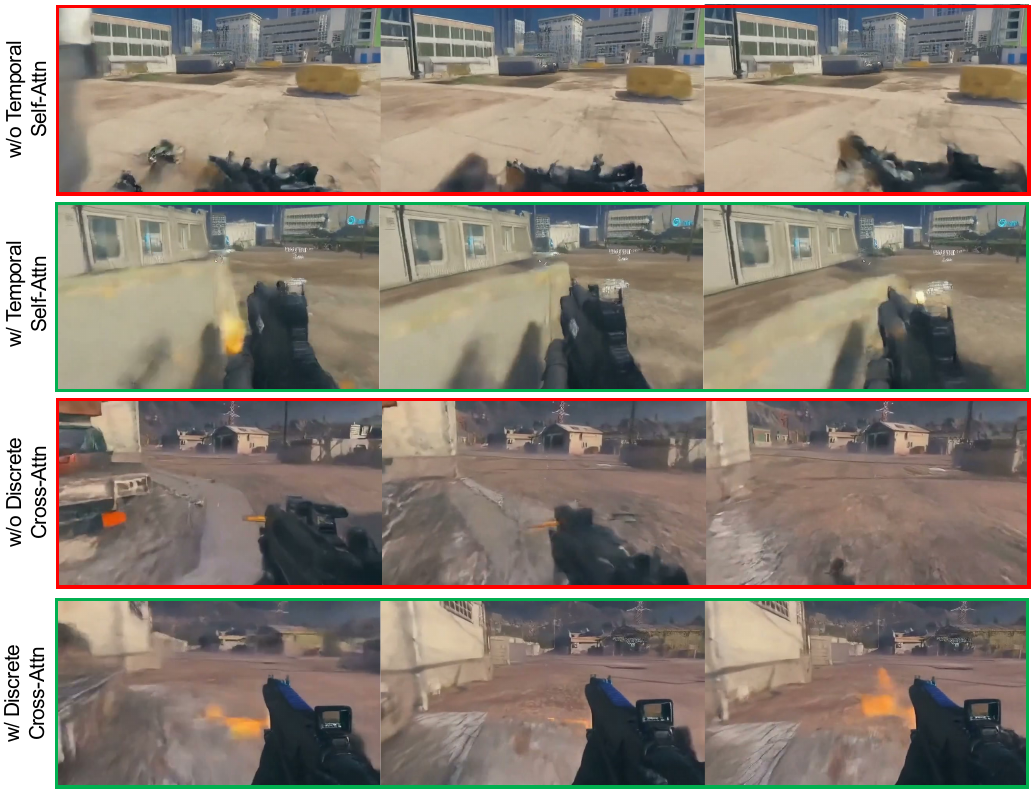}
\end{minipage}
\caption{Qualitative ablation. Left: without spatial selectivity, actions perturb the entire frame (\textcolor{red}{red}); with SCOPE, effects are confined (\textcolor{green}{green}). Right: removing pathway components causes motion degradation or in-scope element loss (\textcolor{red}{red}). Full model preserves both (\textcolor{green}{green}).}
\label{fig:ablation_qual}
\end{figure*}

Removing spatial selectivity causes the most severe degradation: Photometric Smoothness worsens $3.8\times$ (0.198$\to$0.745) and Dynamic Degree drops to 0.521, reproducing global-conditioning failure modes. Removing temporal self-attention collapses Flow Score from 18.24 to 11.60, confirming that dedicated temporal modeling is essential for continuous controls. Removing discrete cross-attention causes effects to leak into out-of-scope regions (Photo.\ 0.198$\to$0.234) while Dynamic Degree remains high (0.846), confirming spatial confinement via visual querying. Without Action-CFG, Dynamic Degree drops to 0.820 and Flow Score to 15.90 due to regression-to-mean attenuation. Figure~\ref{fig:ablation_qual} visualizes these differences: the spatial selectivity variant produces frame-wide distortions under a fire command, whereas the full model confines effects precisely to in-scope regions.

\paragraph{Scalability.}
For training strategy (Table~\ref{tab:ablation_arch}, bottom), performance improves monotonically from Frozen (FVD 775.4) through Two-stage (732.1) to End-to-end (690.3). Performance also scales with data volume and diversity without saturation, suggesting the architecture can benefit from expanded datasets (Appendix~\ref{app:data_ablation}).

\subsection{Generalization to Unseen Scenes}
\label{sec:action_control}

\begin{figure*}[t]
    \centering
    \includegraphics[width=0.95\linewidth]{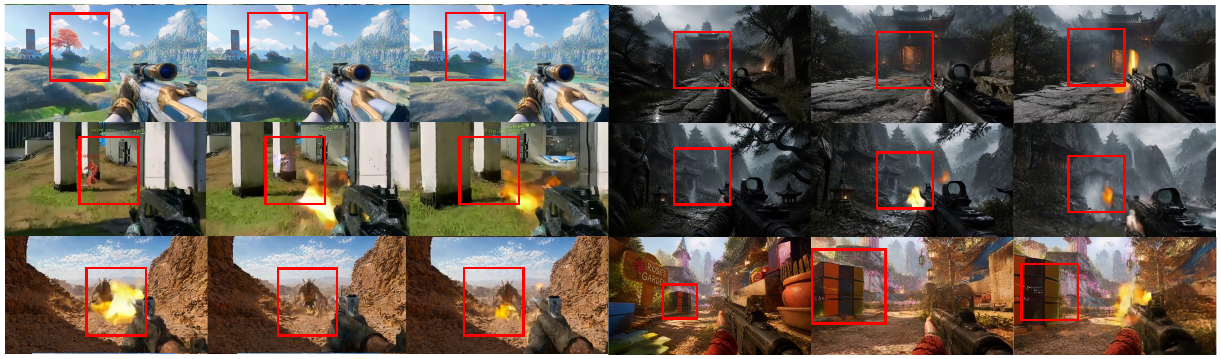}
    \caption{Action controllability on unseen scenes. Left: single and multi-action execution with in-scope effects (\textcolor{red}{red boxes}). Right: action-environment interactions on GPT-image-2 synthesized scenes.}
    \label{fig:action_control}
\end{figure*}

To validate that the model learns general visual-to-action mappings rather than game-specific patterns, we synthesize first-person frames using GPT-image-2~\citep{gptimage2} spanning aesthetics absent from training: stylized open-world, cooperative adventure, mythological action, and sci-fi corridor.

\paragraph{Visual quality.}
We first evaluate whether scope separation and scene stability transfer to unseen aesthetics.

\begin{table}[t]
\centering
\caption{Visual quality on unseen scenes (50 clips per category, first frames from GPT-image-2).}
\label{tab:zero_shot_quality}
\small
\begin{tabular}{lccccc}
\toprule
Scene Style & JEPA$\uparrow$ & LPIPS$\downarrow$ & Flow$\uparrow$ & Photo.$\downarrow$ & Smooth$\uparrow$ \\
\midrule
Stylized open-world & 0.772 & 0.618 & 17.45 & 0.235 & 2.341 \\
Cooperative adventure & 0.758 & 0.632 & 16.89 & 0.251 & 2.298 \\
Mythological action & 0.781 & 0.612 & 17.82 & 0.224 & 2.356 \\
Sci-fi corridor & 0.795 & 0.605 & 18.01 & 0.212 & 2.370 \\
\midrule
Average (unseen) & 0.777 & 0.617 & 17.54 & 0.231 & 2.341 \\
In-distribution (ref.) & 0.806 & 0.601 & 18.24 & 0.198 & 2.383 \\
\bottomrule
\end{tabular}
\end{table}

Table~\ref{tab:zero_shot_quality} shows modest degradation relative to in-distribution performance (JEPA 0.777 vs.\ 0.806, Photo.\ 0.231 vs.\ 0.198). Scenes structurally similar to FPS environments (sci-fi corridors) achieve near-parity. The consistently low Photometric Smoothness across all categories ($\leq$0.251) confirms that scope separation generalizes to novel visual domains.

\paragraph{Action controllability.}
We evaluate tasks at three difficulty levels: single discrete actions, multi-action compositions, and action-environment interactions. For each task, 50 videos are generated from synthesized first frames and assessed via Gemini pre-evaluation with human verification.

\begin{table*}[t]
\centering
\caption{Action controllability on unseen scenes. Completion rate ($N{=}50$ per task).}
\label{tab:action_control}
\small
\resizebox{\textwidth}{!}{%
\begin{tabular}{lccccccccc}
\toprule
\multirow{2}{*}{Method} & \multicolumn{2}{c}{Single Action} & \multicolumn{3}{c}{Multi-Action Composition} & \multicolumn{3}{c}{Action-Environment Interaction} & \multirow{2}{*}{Average} \\
\cmidrule(lr){2-3} \cmidrule(lr){4-6} \cmidrule(lr){7-9}
& Fire & Scope & Scope+Fire & Move+Fire & Switch+Fire & Object & Environment & NPC & \\
\midrule
Matrix-Game 3.0 & 0\% & 0\% & 0\% & 4\% & 0\% & 0\% & 0\% & 0\% & 0.5\% \\
HY-World 1.5 & 4\% & 12\% & 2\% & 36\% & 2\% & 0\% & 6\% & 2\% & 8.0\% \\
LingBot-World (Act) & 82\% & 74\% & 42\% & 18\% & 26\% & 12\% & 32\% & 20\% & 38.3\% \\
\midrule
\SCOPE{} & \textbf{94\%} & \textbf{90\%} & \textbf{82\%} & \textbf{76\%} & \textbf{68\%} & \textbf{46\%} & \textbf{62\%} & \textbf{54\%} & \textbf{71.5\%} \\
\bottomrule
\end{tabular}%
}
\end{table*}

Table~\ref{tab:action_control} shows that our method (71.5\%) outperforms LingBot-World (38.3\%) by $1.9\times$. The gap widens with complexity: single actions 92\% vs.\ 78\%, compositions 75\% vs.\ 29\%, environment interactions 54\% vs.\ 21\%. Matrix-Game 3.0 (0.5\%) and HY-World 1.5 (8.0\%) confirm that global conditioning fails on unseen scenes. Environment effects (62\%) complete more reliably than object deformation (46\%), reflecting the backbone's strength in texture over geometry. Figure~\ref{fig:action_control} shows qualitative examples: left columns show single and multi-action execution with in-scope effects in \textcolor{red}{red boxes}, right columns show NPC and environment interactions on unseen scenes.

\section{Limitations and Future Work}

SCOPE demonstrates effective scope separation and zero-shot transfer to unseen scenes, but its generalization currently covers cross-scene visual transfer and basic action interactions. More complex in-scope behaviors such as multi-step weapon mechanics, item usage, and fine-grained object manipulation remain challenging, due to limited interaction diversity in the training data. The model handles appearance-level responses (fire, smoke, lighting) better than geometric transformations (structural deformation, physics-driven reactions), reflecting the texture bias of the diffusion backbone. Degraded initial frames with extreme blur also cause regression toward the average training appearance.

Despite these limitations, performance scales monotonically with data volume and diversity (Appendix~\ref{app:data_ablation}) without saturation, suggesting that richer interaction data can expand the range of learnable behaviors. In future work, we aim to extend SCOPE to long-horizon, multi-stage task execution, where the world model maintains consistent state across extended gameplay, enabling full game-level control beyond single-clip generation.

\section{Conclusion}

We presented \SCOPE{}, an interactive world model for FPS games that separates in-scope and out-of-scope regions through per-pixel action conditioning. By conditioning each pixel on its local visual content rather than broadcasting global embeddings, the model learns this separation implicitly without segmentation labels. End-to-end training on CrossFPS enables co-adaptation between the pretrained backbone and the SCOPE modules, with performance scaling monotonically with data volume and diversity. From only 69K training clips, the model generalizes zero-shot to unseen game aesthetics. We believe per-pixel conditioning can extend beyond FPS to broader egocentric interactive scenarios. Extending to long-horizon stateful simulation remains an important future direction.

\begin{ack}
We would like to express our sincere gratitude to Ruidong Wang and Murphy Zhao for their tremendous support throughout this project. We are also deeply thankful to Shusen Wang for his invaluable assistance with technical maintenance. 
\end{ack}

{\small
\bibliographystyle{plain}
\bibliography{references}
}


\appendix

\section{CrossFPS Dataset Details}
\label{app:dataset}

This appendix provides complete details on the CrossFPS dataset, organized as follows: Section~\ref{app:overview_stats} presents the dataset overview and per-game statistics; Section~\ref{app:action_format} specifies the action telemetry format; Section~\ref{app:pipeline} describes the data processing pipeline; and Section~\ref{app:captioning} details the text caption generation procedure.

\subsection{Overview and Statistics}
\label{app:overview_stats}

CrossFPS comprises 69,000 five-second clips across seven FPS titles at 20 fps ($480{\times}832$), sourced from two public repositories: NitroGen~\citep{magne2026nitrogen}, which provides gameplay recordings with frame-aligned controller telemetry for the Halo and Call of Duty series, and WorldCam~\citep{nam2026worldcam}, which contributes Xonotic recordings. The dataset is split 95:3:2 into train/val/test sets. Per-game statistics are reported in Table~\ref{tab:dataset_full}.

\begin{table}[h]
\centering
\caption{\textbf{CrossFPS per-game statistics.} All clips are 5 seconds at 20 fps with $480{\times}832$ resolution.}
\label{tab:dataset_full}
\small
\begin{tabular}{lcccc}
\toprule
Game & Total & Train & Val & Test \\
\midrule
Halo Infinite~\citep{haloinfinite} & 32,466 & 30,844 & 973 & 649 \\
Xonotic~\citep{xonotic} & 10,460 & 9,938 & 313 & 209 \\
Call of Duty: Modern Warfare~\citep{codmw} & 8,853 & 8,411 & 265 & 177 \\
Halo~\citep{halo} & 8,227 & 7,817 & 246 & 164 \\
Call of Duty: Warzone~\citep{codwarzone} & 4,818 & 4,578 & 144 & 96 \\
Call of Duty: Modern Warfare III~\citep{codmw3} & 3,662 & 3,480 & 109 & 73 \\
Call of Duty~\citep{cod} & 514 & 489 & 15 & 10 \\
\midrule
\textbf{Total} & \textbf{69,000} & \textbf{65,557} & \textbf{2,065} & \textbf{1,378} \\
\bottomrule
\end{tabular}
\end{table}

\begin{figure*}[h]
    \centering
    \includegraphics[width=\linewidth]{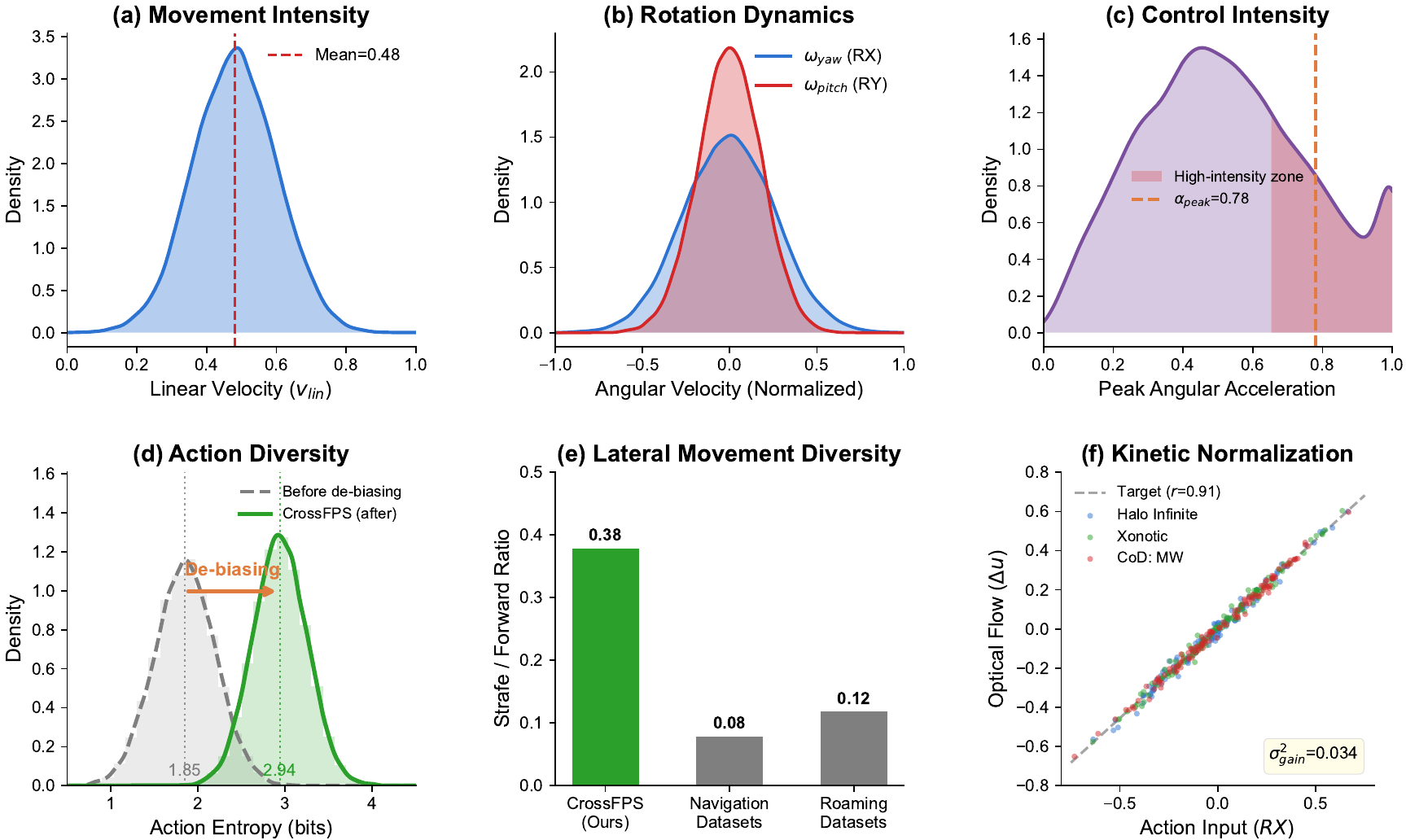}
    \caption{CrossFPS statistics. (a) Linear velocity distribution. (b) Angular velocity for yaw and pitch. (c) Peak angular acceleration with high-intensity zone. (d) Action entropy before and after de-biasing. (e) Strafe-to-forward ratio compared with navigation and roaming datasets. (f) Post-normalization kinetic consistency across three titles ($\sigma^2_{\mathrm{gain}} = 0.034$).}
    \label{fig:dataset_stats}
\end{figure*}

Beyond per-game volume, Table~\ref{tab:detailed_stats} summarizes the kinematic properties and diversity of CrossFPS after processing. All statistics are computed across the 65,557 training clips with continuous signals normalized to $[-1, 1]$. The corresponding distributions are visualized in Figure~\ref{fig:dataset_stats}.

\begin{table}[h]
\centering
\caption{CrossFPS dataset statistics after processing. Continuous signals normalized to $[-1, 1]$.}
\label{tab:detailed_stats}
\small
\setlength{\tabcolsep}{6pt}
\begin{tabular}{lc}
\toprule
Metric & Value (Mean $\pm$ SD) \\
\midrule
\multicolumn{2}{l}{\textit{Action Intensity}} \\
Linear Velocity ($v_{\mathrm{lin}}$) & $0.48 \pm 0.12$ \\
Angular Velocity ($\omega_{\mathrm{ang}}$) & $0.26 \pm 0.18$ \\
Peak Angular Accel. ($\alpha_{\mathrm{peak}}$) & $0.78 \pm 0.14$ \\
Control Smoothness & $0.82 \pm 0.09$ \\
\midrule
\multicolumn{2}{l}{\textit{Distribution}} \\
Action Entropy ($H$) & $2.94 \pm 0.31$ bits \\
Gaze Center-bias Index & $0.42 \pm 0.08$ \\
Strafe-to-Forward Ratio & $0.38 : 1.0$ \\
Discrete Event Density & $14.2\% \pm 3.5\%$ \\
\midrule
\multicolumn{2}{l}{\textit{Cross-Game Consistency}} \\
Optical Flow-Action Corr. ($r$) & $0.91 \pm 0.03$ \\
Inter-game Gain Variance & $0.034$ \\
\bottomrule
\end{tabular}
\end{table}

The high mean linear velocity ($0.48$) results from the $\geq 70\%$ activity filter. The angular velocity ($0.26 \pm 0.18$) covers both precision aiming and rapid flicks, corresponding to approximately $30^\circ$--$60^\circ$/s at 20 fps. The peak angular acceleration ($0.78 \pm 0.14$) confirms abundant high-frequency events (flick shots, 180-degree snap turns) that stress-test scene stability (Figure~\ref{fig:dataset_stats}a--c).

The action entropy of $2.94 \pm 0.31$ bits approaches the theoretical maximum for the discretized 10-dimensional action space, substantially exceeding typical human gameplay entropy. This confirms training without human bias: the model cannot rely on simple temporal priors and must learn physical action-visual mappings. Figure~\ref{fig:dataset_stats}d shows the entropy shift after de-biasing (1.85$\to$2.94 bits). The strafe-to-forward ratio of $0.38:1.0$ is significantly higher than navigation datasets (typically $< 0.1$), introducing motion parallax that forces correct in-scope/out-of-scope separation (Figure~\ref{fig:dataset_stats}e). The gaze center-bias index ($0.42 \pm 0.08$) is lower than typical human play ($0.65$) and professional players ($0.72$), confirming diverse view angles from de-biasing.

The post-normalization gain variance of $0.034$ validates kinetic normalization. Before calibration, the variance across engines exceeds $0.8$ (identical stick displacement produces $10^\circ$ rotation in Halo but $30^\circ$ in Call of Duty), causing gradient conflicts during joint training. After normalization, all titles share a unified action space with $r = 0.91 \pm 0.03$ between input and optical flow (Figure~\ref{fig:dataset_stats}f).

\subsection{Action Telemetry Format}
\label{app:action_format}

Each clip is paired with per-frame 10-dimensional controller telemetry organized into four functional groups (Table~\ref{tab:action_format}): Movement (4 continuous axes from the left analog stick), Camera (2 continuous axes from the right analog stick), Combat (3 discrete buttons), and Utility (3 discrete buttons). The continuous signals capture analog intensity (e.g., partial stick deflection for slow movement), while discrete signals are binary indicators sampled at each frame.

\begin{table}[h]
\centering
\caption{\textbf{Action telemetry format.} 10 dimensions organized into 4 functional groups.}
\label{tab:action_format}
\small
\begin{tabular}{llll}
\toprule
Group & Input & Type & Description \\
\midrule
\multirow{2}{*}{\textsc{Movement}} & LX & continuous & Move left / right \\
& LY & continuous & Move forward / back \\
\midrule
\multirow{2}{*}{\textsc{Camera}} & RX & continuous & Turn left / right \\
& RY & continuous & Look up / down \\
\midrule
\multirow{3}{*}{\textsc{Combat}} & RT & discrete & Fire \\
& LT & discrete & Aim down sights (ADS) \\
& R3 & discrete & Melee \\
\midrule
\multirow{3}{*}{\textsc{Utility}} & A & discrete & Jump \\
& X & discrete & Reload \\
& Y & discrete & Switch weapon \\
\bottomrule
\end{tabular}
\end{table}

\subsection{Data Processing Pipeline}
\label{app:pipeline}

A key design goal of CrossFPS is to eliminate human bias---the tendency of skilled players to execute stereotyped action patterns (e.g., always firing at highlighted enemies or crouching behind cover). This ensures that SCOPE learns authentic \emph{physical action-visual mappings} rather than memorizing game strategies. To achieve this, we process the raw gameplay recordings through a rigorous pipeline designed to enforce diversity, balance, and cross-game consistency, structured into four primary phases:

\paragraph{Spatial-Temporal Formatting.} 
We first extract the active game area by cropping out streaming overlays and UI borders. Videos are split at scene transitions (e.g., death or loading screens) using frame-level visual similarity to ensure continuous gameplay. We then segment the recordings into non-overlapping 5-second windows and normalize the frame rate to a uniform 20 fps via temporal subsampling (for 60 fps sources) or interpolation (for 30 fps sources). Finally, game-specific UI elements (like chat boxes) are cropped out, and all clips are resized to $480 \times 832$ to maintain a 16:9 aspect ratio.

\paragraph{Quality Filtering and Action Balancing.} 
Human gameplay inherently suffers from a long-tail distribution, heavily skewed toward low-intensity states (e.g., straight-line running). We first apply an activity filter (left-stick active $\geq 70\%$) to remove idle clips. To ensure adequate coverage of high-intensity dynamics, we compute the action entropy $H_i = -\sum_k p_k \log p_k$ and peak camera velocity for each clip. High-intensity clips (the top 15\%, featuring rapid 180-degree flicks or jump chains) are oversampled by $3\times$. This step prevents the model from collapsing into generating only smooth, low-motion sequences.

\paragraph{Visual-Action De-biasing.} 
To force the model to learn raw physics rather than strategic priors, we explicitly retain "inefficient" or counter-intuitive actions (e.g., firing at an empty sky, sprinting into walls). We identify these clips by computing the mutual information between the visual features from a pre-trained scene classifier and the discrete action sequences. Clips with the lowest mutual information (the bottom 20\%) are flagged as "de-biased" samples and forcefully included in the training set. This teaches SCOPE that actions reliably trigger corresponding visual changes regardless of their strategic utility.

\paragraph{Cross-Game Kinetic Normalization.} 
Different game engines map analog stick displacements to vastly different camera rotation speeds (e.g., identical stick displacement produces a $10^\circ$ rotation in \textit{Halo} but $30^\circ$ in \textit{Call of Duty}). To resolve the resulting gradient conflicts during multi-game joint training, we apply optical flow-based gain calibration. For each clip, we extract the mean pixel displacement $(\Delta u, \Delta v)$ caused by camera rotation and fit a linear gain model ($\Delta u \approx g_x \cdot RX$). Camera signals are rescaled by $RX_{\mathrm{norm}} = RX \cdot (\bar{g}_x / g_x)$, where $\bar{g}_x$ is the dataset-wide mean gain. For static or highly occluded scenes where optical flow fails, we apply a 95th-percentile fallback normalization. Additionally, inverted axes (e.g., in \textit{Xonotic}) are negated to establish a unified directional convention.

\vspace{0.5em}
Prior to training, all 65,557 training clips passed a comprehensive integrity check (validating video readability, first-frame decodability, frame count, resolution, and action file completeness) with a 100\% pass rate.

\subsection{Text Caption Generation}
\label{app:captioning}

\begin{figure*}[h]
    \centering
    \includegraphics[width=\linewidth]{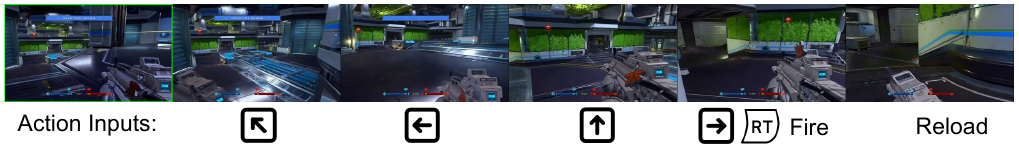}
    \caption{\textbf{Halo Infinite example.} The first frame (highlighted with a \textcolor{green}{green box}) is used for caption generation and as the image-to-video condition. The action input sequence shows forward movement with a rightward camera sweep followed by ADS activation. Caption: ``A futuristic sci-fi indoor arena with multi-level platforms, neon blue lighting, green vegetation behind glass walls, and metallic surfaces during an overtime round. The player is holding a large rifle in a ready stance, with the score tied 0-0 at 4:57 remaining, and a crosshair centered on the screen.''}
    \label{fig:caption_halo}
\end{figure*}

\begin{figure*}[h]
    \centering
    \includegraphics[width=\linewidth]{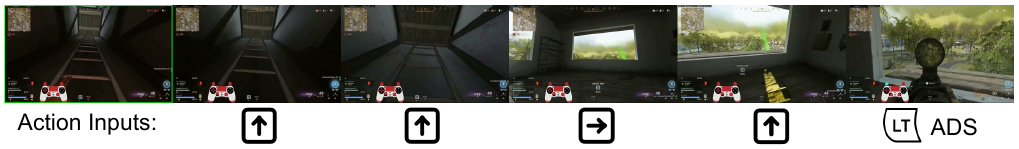}
    \caption{\textbf{Call of Duty: Warzone example.} The first frame (highlighted with a \textcolor{green}{green box}) is used for caption generation and as the image-to-video condition. The action input sequence shows a leftward camera rotation transitioning to forward movement with simultaneous fire and reload events. Caption: ``A dark narrow stairwell inside a building in Caldera Capital City, with a wooden ladder leading upward through a dimly lit vertical shaft. The player is climbing the ladder while holding a weapon with 48 rounds, a minimap and kill feed visible on the HUD, and a controller overlay displayed at the bottom center of the screen.''}
    \label{fig:caption_cod}
\end{figure*}

\begin{figure*}[h]
    \centering
    \includegraphics[width=\linewidth]{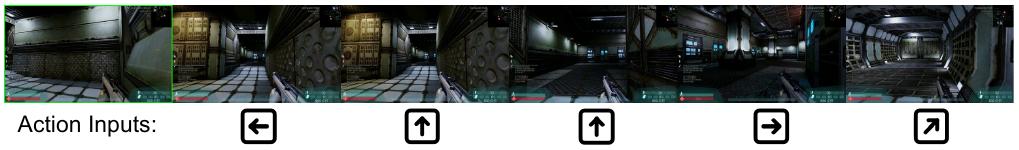}
    \caption{\textbf{Xonotic example.} The first frame (highlighted with a \textcolor{green}{green box}) is used for caption generation and as the image-to-video condition. The action input sequence shows leftward movement combined with forward camera motion, rightward sweep, and a diagonal turn. Caption: ``A dark military-industrial interior room labeled `Computer Room' with large metal panel walls featuring riveted circular patterns, grid-patterned flooring, and dim greenish lighting. The player is holding a tan assault rifle at hip level, with full health at 100, 30 rounds in the magazine, 800 points displayed, and running at 125 fps.''}
    \label{fig:caption_xonotic}
\end{figure*}

To provide text conditioning during training, we generate scene descriptions for the first frame of every clip using Gemini~\citep{team2023gemini}. Each caption follows a standardized two-sentence format:
\begin{itemize}
    \item \textbf{Sentence 1} describes the environment: setting, lighting, architecture, and atmosphere.
    \item \textbf{Sentence 2} describes the player state and salient visual elements: weapon type, HUD indicators, nearby objects, and game-specific UI.
\end{itemize}
This structured format ensures consistent conditioning signals across all games while preserving scene-specific details. Representative examples from each game are shown in Figure~\ref{fig:caption_halo}--\ref{fig:caption_xonotic}, where the first frame used for caption generation and as the image-to-video condition is highlighted with a \textcolor{green}{green box}, and the corresponding frame-aligned action inputs are displayed below the frame sequence.

\section{Implementation Details}
\label{app:implementation}

The backbone model is Wan2.2-TI2V-5B~\citep{wan} with 30 transformer layers, hidden dimension 3072, 24 attention heads, patch size $[1,2,2]$, and FFN dimension 14336. The text encoder is UMT5-XXL producing 4096-dimensional embeddings. The VAE uses 8$\times$ spatial compression and 4$\times$ temporal compression. Each SCOPE module contains: a fusion MLP for continuous control processing, a cross-attention block for discrete event processing, and temporal RoPE embeddings. All output projections are zero-initialized so that modules produce no perturbation at initialization. The entire model (backbone + SCOPE modules) is trained end-to-end using AdamW with learning rate $10^{-5}$, bfloat16 precision, gradient checkpointing, and batch size 1 per GPU across 8 NVIDIA GPUs for 500 epochs. We use the Accelerate library for distributed training with DDP.

\subsection{Training and Inference Pseudocode}

The complete training and inference procedures are given in Algorithm~\ref{alg:training} and Algorithm~\ref{alg:sampling}.

\begin{algorithm}[h]
  \caption{SCOPE training} \label{alg:training}
  \begin{algorithmic}[1]
    \REQUIRE $p_{\mathrm{drop}}$: action dropout probability; $\mathbf{a}_{\mathrm{null}}$: learnable null embedding
    \REPEAT
      \STATE $(\mathbf{V}, \mathbf{a}_c, \mathbf{a}_d, \mathbf{c}_{\mathrm{text}}) \sim \mathcal{D}$ \COMMENT{Sample from CrossFPS}
      \STATE $\mathbf{z}_0 \gets \mathrm{VAE}_{\mathrm{enc}}(\mathbf{V})$; \quad condition on first-frame latent $\mathbf{z}_0^{(1)}$
      \STATE $(\mathbf{a}_c, \mathbf{a}_d) \gets \mathbf{a}_{\mathrm{null}}$ with probability $p_{\mathrm{drop}}$ \COMMENT{Action dropout for CFG}
      \STATE $t \sim \mathcal{U}(0, 1)$; \quad $\boldsymbol{\epsilon} \sim \mathcal{N}(\mathbf{0}, \mathbf{I})$; \quad $\mathbf{z}_t \gets (1-t)\mathbf{z}_0 + t\boldsymbol{\epsilon}$
      \STATE $\mathbf{x} \gets \mathrm{Patchify}(\mathbf{z}_t)$
      \FOR{$l = 1, \dotsc, L$}
        \STATE $\mathbf{x} \gets \mathrm{DiTBlock}_l(\mathbf{x}, \mathbf{c}_{\mathrm{text}}, t)$ \COMMENT{Standard DiT: self-attn + text cross-attn}
        \STATE $\hat{\mathbf{x}} \gets \mathrm{Reshape}(\mathbf{x})$ to per-pixel temporal sequences \COMMENT{Spatial selectivity}
        \STATE $\Delta \mathbf{x}_c \gets \mathrm{SelfAttn}(\mathrm{MLP}_{\mathrm{fuse}}([\hat{\mathbf{x}}; \mathbf{a}_c]))$ \COMMENT{Continuous pathway}
        \STATE $\Delta \mathbf{x}_d \gets \mathrm{CrossAttn}(Q{=}\hat{\mathbf{x}},\, K{=}V{=}\mathrm{MLP}_{\mathrm{embed}}(\mathbf{a}_d))$ \COMMENT{Discrete pathway}
        \STATE $\mathbf{x} \gets \mathrm{Reshape}(\hat{\mathbf{x}} + \Delta \mathbf{x}_c + \Delta \mathbf{x}_d)$; \quad $\mathbf{x} \gets \mathrm{FFN}_l(\mathbf{x})$
      \ENDFOR
      \STATE Take gradient step on $w(t)\left\| \mathbf{v}_\theta - (\boldsymbol{\epsilon} - \mathbf{z}_0) \right\|^2$
    \UNTIL{converged}
  \end{algorithmic}
\end{algorithm}

\begin{algorithm}[h]
  \caption{SCOPE inference with Action-CFG} \label{alg:sampling}
  \begin{algorithmic}[1]
    \REQUIRE $\lambda$: guidance strength; $\mathbf{I}_1$: first frame; $(\mathbf{a}_c, \mathbf{a}_d)$: action sequence; $t_1{>}\cdots{>}t_S$: schedule
    \STATE $\mathbf{z}_0^{(1)} \gets \mathrm{VAE}_{\mathrm{enc}}(\mathbf{I}_1)$; \quad $\mathbf{z}_{t_1} \sim \mathcal{N}(\mathbf{0}, \mathbf{I})$
    \FOR{$s = 1, \dotsc, S$}
      \STATE $\mathbf{v}_{\mathrm{cond}} \gets \mathbf{v}_\theta(\mathbf{z}_{t_s}, t_s, \mathbf{c}, \mathbf{a}_c, \mathbf{a}_d)$ \COMMENT{Conditional forward pass}
      \STATE $\mathbf{v}_{\mathrm{uncond}} \gets \mathbf{v}_\theta(\mathbf{z}_{t_s}, t_s, \mathbf{c}, \mathbf{a}_{\mathrm{null}})$ \COMMENT{Unconditional forward pass}
      \STATE $\hat{\mathbf{v}} \gets \mathbf{v}_{\mathrm{uncond}} + \lambda (\mathbf{v}_{\mathrm{cond}} - \mathbf{v}_{\mathrm{uncond}})$ \COMMENT{Action-CFG}
      \STATE $\mathbf{z}_{t_{s+1}} \gets \mathrm{ODEStep}(\mathbf{z}_{t_s}, \hat{\mathbf{v}}, t_s \to t_{s+1})$
    \ENDFOR
    \STATE \textbf{return} $\mathrm{VAE}_{\mathrm{dec}}(\mathbf{z}_{t_{S+1}})$
  \end{algorithmic}
\end{algorithm}

\section{Evaluation Metrics}
\label{app:metrics}

We evaluate the generated videos along three primary axes using eight metrics. All metrics are computed on the CrossFPS test set (1,378 clips) at a $480 \times 832$ resolution. A comprehensive summary of the evaluation metrics is provided in Table~\ref{tab:metrics_summary}.

\begin{table}[h]
  \centering
  \caption{Summary of Evaluation Metrics for Reactive Game World Models.}
  \label{tab:metrics_summary}
  \resizebox{\textwidth}{!}{
  \begin{tabular}{llll}
    \toprule
    \textbf{Evaluation Axis} & \textbf{Metric} & \textbf{Reference} & \textbf{Primary Focus} \\
    \midrule
    \multirow{2}{*}{Action Responsiveness} 
    & Dynamic Degree & \citep{huang2024vbench} & Quantifies overall video activity and motion magnitude. \\
    & Flow Score & \citep{liu2024evalcrafter} & Measures average optical flow displacement between frames. \\
    \midrule
    \multirow{2}{*}{Spatial Stability} 
    & Photometric Smoothness & \citep{duan2025worldscore} & Evaluates pixel-level color consistency via backward-warping. \\
    & Depth Accuracy & \citep{shang2026worldarena} & Assesses 3D geometric consistency using depth reprojection. \\
    \midrule
    \multirow{4}{*}{Visual Quality} 
    & JEPA Similarity & \citep{bardes2024vjepa,luo2024beyond} & Captures high-level semantic and physical structural fidelity. \\
    & FVD & \citep{unterthiner2018towards} & Measures the realism of the generated spatiotemporal distribution. \\
    & LPIPS & \citep{zhang2018unreasonable} & Evaluates per-frame perceptual distortion and image clarity. \\
    & Motion Smoothness & \citep{duan2025worldscore,zhang2024vfimamba} & Analyzes flow acceleration to penalize abrupt motion jitter. \\
    \bottomrule
  \end{tabular}
  }
\end{table}

\subsection{Action Responsiveness}

This axis measures whether generated videos exhibit genuine dynamic changes in response to sequential action inputs, heavily penalizing models that produce static or unresponsive outputs.

\paragraph{Dynamic Degree~\citep{huang2024vbench}.} We extract spatiotemporal features from the generated video sequence to quantify the magnitude of inter-frame dynamic variation. By analyzing the intensity of object motion and global scene changes across frames, this metric produces a holistic score reflecting the overall activity level.

\paragraph{Flow Score~\citep{liu2024evalcrafter}.} We compute the average optical flow magnitude between consecutive frames. Let $F_t(x,y) \in \mathbb{R}^2$ denote the estimated optical flow vector at spatial coordinate $(x,y)$ from frame $I_t$ to $I_{t+1}$. The Flow Score across $T$ frames with spatial dimensions $H \times W$ is defined as:
$$FS = \frac{1}{(T-1)HW} \sum_{t=1}^{T-1} \sum_{x=1}^{W} \sum_{y=1}^{H} \|F_t(x,y)\|_2$$
Higher scores indicate a larger motion amplitude; conversely, near-zero scores typically signal generation failure (i.e., frozen frames).

\subsection{Spatial Stability}

This axis measures whether generated videos maintain a consistent 3D geometric structure over time, directly penalizing spatial collapse, warping, or unauthorized object deformation.

\paragraph{Photometric Smoothness~\citep{duan2025worldscore}.} We evaluate pixel-level color consistency between adjacent frames. Using estimated depth maps and optical flow, we backward-warp pixels from the next frame $I_{t+1}$ to the current frame $I_t$'s viewpoint to obtain the reconstructed frame $\tilde{I}_t$. The photometric error is measured as the average $L_1$ distance:
$$PS = \frac{1}{(T-1)HW} \sum_{t=1}^{T-1} \sum_{x=1}^{W} \sum_{y=1}^{H} \|I_t(x,y) - \tilde{I}_t(x,y)\|_1$$
Lower errors indicate highly stable photometric behavior devoid of flickering or texture artifacts.

\paragraph{Depth Accuracy~\citep{shang2026worldarena}.} We predict depth maps $D_t$ for all generated frames using a pre-trained monocular depth estimator. Given the camera pose transformation $T_{t \to t+1}$, we reproject the 3D point cloud from the previous frame into the current frame to compute the reprojected depth $\tilde{D}_{t+1}$. We calculate the scale-invariant relative error between $\tilde{D}_{t+1}$ and the directly estimated $D_{t+1}$. Higher accuracy confirms that the model rigidly maintains correct 3D scene geometry.

\subsection{Visual Quality}

This axis rigorously evaluates image clarity, perceptual realism, semantic fidelity, and motion coherence.

\paragraph{JEPA Similarity~\citep{bardes2024vjepa,luo2024beyond}.} We extract feature vectors from both generated videos $V$ and ground-truth reference videos $\hat{V}$ using a pre-trained Joint Embedding Predictive Architecture (V-JEPA). Because V-JEPA captures high-level semantic content without relying on strict pixel-level reconstruction, we compute the cosine similarity in its feature space $\phi$:
$$S_{\text{JEPA}} = \frac{\phi(V) \cdot \phi(\hat{V})}{\|\phi(V)\|_2 \|\phi(\hat{V})\|_2}$$
Higher similarity demonstrates that the generated content preserves the semantic and physical structure of the reference sequence.

\paragraph{FVD~\citep{unterthiner2018towards}.} We extract spatiotemporal features from both generated and real video collections using a pre-trained I3D network. Modeling the feature distributions of the real data and generated data as multivariate Gaussians $\mathcal{N}(\mu_r, \Sigma_r)$ and $\mathcal{N}(\mu_g, \Sigma_g)$, the Fr\'{e}chet Video Distance is calculated as:
$$FVD = \|\mu_r - \mu_g\|^2_2 + \text{Tr}(\Sigma_r + \Sigma_g - 2(\Sigma_r\Sigma_g)^{1/2})$$
A lower FVD indicates that the generated spatiotemporal distribution more closely matches the real data distribution.

\paragraph{LPIPS~\citep{zhang2018unreasonable}.} We measure per-frame perceptual distortion by passing generated frames $\hat{x}$ and reference frames $x$ through a pre-trained VGG network. We compute the weighted $L_2$ distance between the normalized intermediate feature maps $\hat{y}^l$ and $y^l$ at each layer $l$:
$$LPIPS(x, \hat{x}) = \sum_l \frac{1}{H_l W_l} \sum_{i,j} \|w_l \odot (\hat{y}^l_{i,j} - y^l_{i,j})\|^2_2$$
LPIPS correlates much more closely with human perception of blur and structural artifacts than traditional pixel-level metrics such as PSNR or SSIM.

\paragraph{Motion Smoothness~\citep{duan2025worldscore,zhang2024vfimamba}.} We analyze the temporal variation of optical flow fields by isolating the acceleration (the second-order derivative) of flow vectors across consecutive frame triplets. Let $A_t = F_{t+1} - F_t$ represent the change in optical flow. We penalize large magnitudes of $A_t$, as abrupt changes in motion trajectory or velocity manifest as visual jitter or stuttering. High motion smoothness scores indicate physically plausible, continuous inertial motion.

\section{Scalability Details}
\label{app:data_ablation}

This section provides full numerical results for the scalability analysis discussed in Section~\ref{sec:ablation}, covering both training strategy comparisons and data scale/diversity ablations.

\paragraph{Training strategy details.} Table~\ref{tab:ablation_arch} (main paper) reports three training regimes applied to the full 65K dataset: (1)~\emph{Frozen backbone}---only the 30 SCOPE modules are trained while all pretrained parameters are fixed; (2)~\emph{Two-stage}---SCOPE modules are first trained with a frozen backbone, then the entire model is fine-tuned jointly; (3)~\emph{End-to-end}---all parameters are trained from the start. The Frozen variant's JEPA of 0.724 and Photometric Smoothness of 0.264 demonstrate that the SCOPE module functions effectively as a plug-and-play adapter. The Two-stage variant (JEPA 0.761) shows that partial backbone adaptation captures mid-level action-visual correlations. Full end-to-end training (JEPA 0.806) yields the strongest results by enabling deep co-adaptation between the SCOPE modules and the backbone's internal representations, particularly for Flow Score (15.57$\to$17.13$\to$18.24). Notably, even the Frozen variant's Photometric Smoothness (0.264) remains far superior to the ``w/o Spatial Selectivity'' ablation (0.745), confirming that the per-pixel conditioning design itself---independent of backbone adaptation---drives out-of-scope stability.

\paragraph{Data scale and diversity configurations.} We study how data volume and source diversity jointly influence model quality by constructing training subsets at five scales with controlled diversity levels. Starting from Halo series data (the largest single source in CrossFPS), we progressively include additional game families:
\begin{itemize}
    \item \textbf{1K} --- 1,000 clips randomly sampled from Halo Infinite (1 title).
    \item \textbf{5K} --- 5,000 clips from Halo Infinite + Halo MCC (2 titles, same series).
    \item \textbf{10K} --- 10,000 clips from Halo series + Call of Duty: Modern Warfare (3 titles, 2 series).
    \item \textbf{30K} --- 30,000 clips from Halo series + CoD series + Xonotic (6 titles, 3 series).
    \item \textbf{65K (full)} --- All 65,557 training clips across 7 titles (3 series).
\end{itemize}
At each scale, clips are randomly sampled from all available titles with balanced per-game ratios (capped at available clips per title). All configurations use single-stage training at $480{\times}832$.

\begin{table}[h]
\centering
\caption{Data scale and diversity ablation. Scale: number of training clips; Titles: number of distinct games; Series: number of distinct game franchises. All trained at $480{\times}832$ with single-stage strategy.}
\label{tab:data_ablation_full}
\small
\setlength{\tabcolsep}{4pt}
\begin{tabular}{rcclccc}
\toprule
Scale & Titles & Series & Source Composition & FVD$\downarrow$ & LPIPS$\downarrow$ & Flow$\uparrow$ \\
\midrule
1K & 1 & 1 & Halo Infinite & 478.20 & 0.545 & 17.64 \\
5K & 2 & 1 & Halo Infinite + Halo MCC & 603.91 & 0.592 & 16.67 \\
10K & 3 & 2 & Halo $\times$2 + CoD:MW & 1017.82 & 0.745 & 11.69 \\
30K & 6 & 3 & Halo $\times$2 + CoD $\times$3 + Xonotic & 799.70 & 0.666 & 16.84 \\
65K & 7 & 3 & Full CrossFPS & \textbf{690.30} & \textbf{0.601} & \textbf{18.24} \\
\bottomrule
\end{tabular}
\end{table}

\paragraph{Training strategy comparison at each scale.} We additionally compare single-stage training (directly at $480{\times}832$) with progressive training ($248{\times}448 \to 480{\times}832$) at each data scale.

\begin{table}[h]
\centering
\caption{\textbf{Training strategy comparison across data scales.} Progressive: low-resolution warm-up followed by high-resolution fine-tuning; Single-stage: trained directly at full resolution.}
\label{tab:training_strategy_full}
\small
\setlength{\tabcolsep}{5pt}
\begin{tabular}{llccc}
\toprule
Scale & Strategy & FVD$\downarrow$ & LPIPS$\downarrow$ & Flow$\uparrow$ \\
\midrule
1K (Halo only) & Progressive & 612.35 & 0.621 & 15.82 \\
1K (Halo only) & Single-stage & 485.42 & 0.554 & 17.95 \\
\midrule
5K (Halo series) & Progressive & 705.18 & 0.645 & 15.30 \\
5K (Halo series) & Single-stage & 621.76 & 0.589 & 16.54 \\
\midrule
10K (2 series) & Progressive & 845.21 & 0.682 & 14.85 \\
10K (2 series) & Single-stage & 1032.55 & 0.741 & 11.23 \\
\midrule
30K (3 series) & Progressive & 781.45 & 0.659 & 16.50 \\
30K (3 series) & Single-stage & 798.30 & 0.645 & 16.62 \\
\midrule
65K (full) & Progressive & 756.28 & 0.652 & 17.12 \\
65K (full) & Single-stage & \textbf{690.30} & \textbf{0.601} & \textbf{18.24} \\
\bottomrule
\end{tabular}
\end{table}

\paragraph{Key observations.} (1) At small scale with limited diversity (1K--5K, single series), single-stage training works well as the domain is homogeneous. (2) At intermediate scale with moderate diversity (10K, 2 series), cross-domain interference destabilizes single-stage training (FVD 1033 vs. 845 for progressive), as visually distinct game assets create conflicting gradients. (3) At full scale with maximum diversity (65K, 7 titles), sufficient multi-source variety provides natural regularization, and single-stage training surpasses progressive across all metrics. These results motivate our final design: single-stage training on the full multi-game dataset.


\newpage

\end{document}